\documentclass[runningheads]{llncs}
\usepackage[T1]{fontenc}
\usepackage{graphicx,verbatim}
\usepackage{amsmath}
\usepackage{amsfonts}
\begin{document}

\title{ExpOS: Explainable Open-Surgery Skills Assessment Using 3D Hand Reconstruction}
%

\author{Roi Papo\inst{1}\and
Idan Smoller\inst{1} \and
Shlomi Laufer\inst{1}}
\authorrunning{Roi Papo et al.}
%
\institute{Faculty of Data and Decision Sciences, Technion – Israel Institute of
Technology, Haifa, 3200003, Israel}

\maketitle              

\begin{abstract}
Timely and transparent feedback is essential for effective surgical training, yet current assessment remains dependent on expert observation, limiting scalability and opportunities for autonomous practice. We present \textbf{ExpOS}, an explainable framework for data-driven assessment of open-surgery skills designed to enable automatic, feedback-oriented evaluation. Rather than relying on expert-defined metrics, ExpOS learns discriminative temporal patterns directly from motion data and identifies the segments and behaviors most predictive of skill level.
We trained and evaluated the method on 221 videos of medical students performing three open-surgery tasks.
Hand poses and tool detections were extracted from each frame to derive kinematic descriptors and global motion statistics. Spatiotemporal hand–tool dynamics were modeled using a temporal convolutional backbone with attention-based pooling to generate frame-level importance maps. These representations were fused with global motion statistics to predict skill level and to provide interpretable feedback. 
ExpOS provides multi-level explainability by identifying when informative events occur through attention weights and which motion characteristics most influence predictions through global feature analysis. Across tasks, the framework achieved strong correlation with expert ratings, with best performance on fascial closure (r = 0.778, $R^2$ = 0.74). These results demonstrate that combining weakly-supervised temporal importance learning with interpretable motion statistics enables scalable and actionable surgical skill assessment.

Code available at : \url{https://github.com/RoiPapo/ExpOS}

\end{abstract}
\begin{figure}[!htb]
        \centering
        \includegraphics[width=0.99\linewidth]{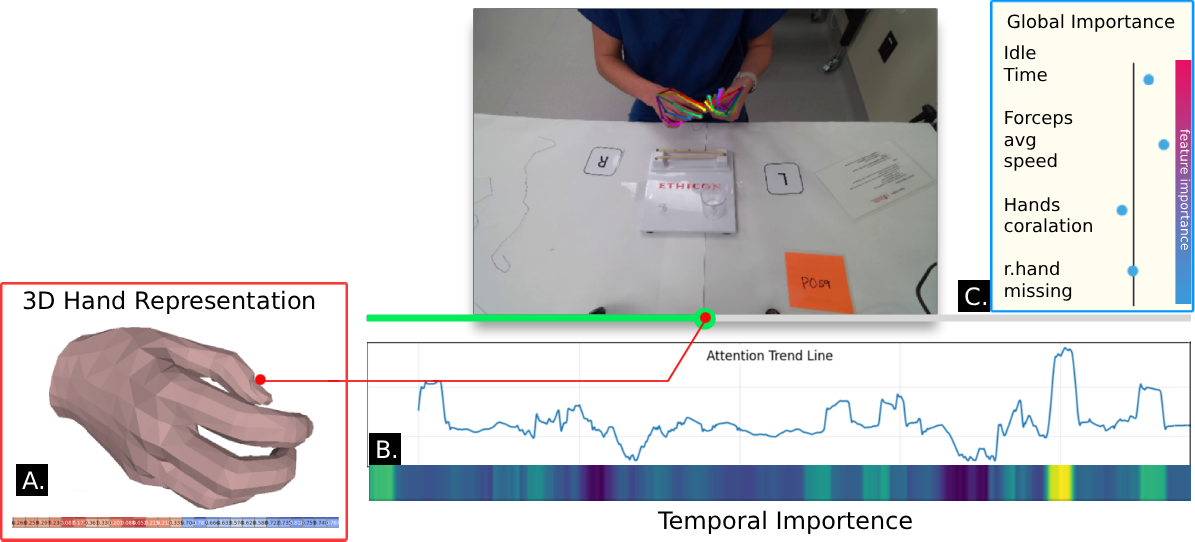}
        \caption{
            Overview of ExpOS explainability outputs: 
            (A) 3D hand pose, 
            (B) temporal attention highlights key video frames, and 
            (C) SHAP-based global feature importance reveals which motion statistics most influenced the skill score.
        }
        \label{fig:explainability_overview}
\end{figure}

\section{Introduction}\label{sec1}
Scalable and objective surgical skill assessment is essential for medical education, particularly when expert supervision is limited. Traditional evaluation using structured rating systems such as OSATS~\cite{martin1997objective} is resource-intensive and subjective~\cite{vedula2017objective,funke2019video}, limiting opportunities for autonomous training. Recent deep learning approaches can predict overall skill levels from motion or video data~\cite{lam2022machine}, but typically provide only global scores without explaining which temporal segments or behavioral patterns drive the prediction. Conversely, methods based on expert-defined metrics offer interpretability but constrain discovery to predefined criteria~\cite{bkheet2023hand}.

We propose \textbf{ExpOS}, an explainable framework for data-driven surgical skill assessment that enables automatic, feedback-oriented evaluation. 3D hand trajectories and kinematic descriptors are modeled using an MS-TCN++ backbone~\cite{li2020ms} to capture long-range temporal dependencies at frame-level resolution. A multi-head attention pooling module assigns importance weights to temporal segments, providing explicit localization of performance-critical moments. In parallel, global motion statistics are fused with learned temporal features, allowing feature-level attribution of interpretable metrics. This design delivers multi-level explainability by identifying both \textit{when} informative events occur and \textit{which} motion characteristics most influence skill prediction. Experiments on three open-surgery tasks demonstrate strong correlation with expert ratings while generating actionable feedback for self-training.

\section{Related Work}

Early computational approaches relied on kinematic features extracted from robotic systems or motion sensors, such as path length, smoothness, and idle time, which were used to classify skill levels~\cite{tao2012sparse,gao2014jigsaws,d2015idle}. These methods depend on expert-defined metrics and specialized hardware. With advances in computer vision, video-based approaches replaced sensor-dependent systems, enabling end-to-end learning from RGB recordings~\cite{yanik2022deep,goldbraikh2022video}. Deep architectures including TCNs, 3D-CNNs, and transformers achieve strong agreement with expert ratings~\cite{funke2019video,hoffmann2024aixsuture}, but typically provide only global skill predictions without temporal localization or interpretable feedback.

To improve transparency, attention mechanisms and saliency-based methods have been introduced to highlight informative temporal segments~\cite{ismail2019accurate,tjoa2020survey}. However, existing approaches ~\cite{anastasiou2023keep,bkheet2023hand} either focus solely on temporal localization without feature-level attribution, or rely on predefined statistical criteria that limit data-driven discovery.

In contrast, our framework jointly performs weakly-supervised temporal importance estimation and feature-level attribution by combining an MS-TCN++ backbone with multi-head attention pooling and interpretable motion statistics. This hybrid design enables both discovery of discriminative temporal patterns and quantification of clinically meaningful motion characteristics, providing multi-level explainability for autonomous training.

\section{Methods}
\subsection{Dataset}
This study analyzed 77 first- and second-year surgical residents from a Midwestern academic hospital collected over three years. Residents participated in an annual simulation series involving standardized basic surgical tasks. The study was approved by the IRB committee. Participation was voluntary, with informed consent obtained from all participants. One week before the simulations, residents received expert demonstration videos. During the sessions, they followed written task instructions and were evaluated by faculty on a 1–10 scale with post-task feedback.

\begin{figure}[!htb]
    \centering
    \includegraphics[width=\linewidth]{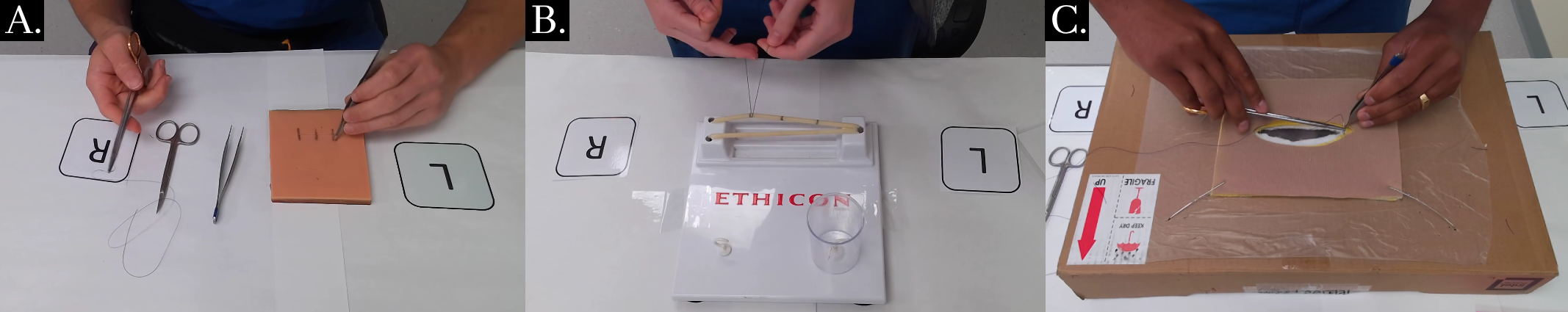}
    \caption{Overview of the procedures executed by the surgical residents A. Suturing, B. Knot Tying, C. Facial closure.}
    \label{fig:simulators_and_hist}
\end{figure}

Thress surgical tasks were performed using three distinct simulators: a suture pad (n=77), a knot-tying station (n=72), and a fascial closure model (n=72) (Figure ~\ref{fig:simulators_and_hist}). Task execution was recorded with a 4K RGB camera (Azure Kinect) operating at 30 frames per second.

A YOLO~\cite{Jocher_Ultralytics_YOLO_2023} model was trained to detect surgical tools (forceps, needle driver, scissors, simulator) using 731 annotated frames with augmentation to ensure robustness. For hand detection, we employed RoHan’s~\cite{Papo2025RoHan} pseudo-label self-training to refine a YOLO-based detector, which was then integrated into WiLoR~\cite{potamias2024wilor} for 3D hand reconstruction. This setup enabled precise 3D tracking of 21 joints per hand throughout each procedure.

\paragraph{Frame-Level Features}
 Each frame was  encoded using both bounding box information and reconstructed 3D hand keypoints, providing a compact yet informative representation of hand–tool interactions. Specifically, six detected objects (left hand, right hand, forceps, needle driver, scissors, and simulator) were represented by their bounding boxes $(X_c, Y_c, W, H)$, flattened into a vector of size $6 \times 4$. In parallel, each detected hand was represented by 21 ordered 3D keypoints, yielding a flattened vector of size $2 \times 21 \times 3$. To further improve temporal consistency, short gaps of up to 10 frames were interpolated linearly, and the resulting trajectories were smoothed using a Savitzky–Golay filter~\cite{Savitzky_filter}. Concatenating the bounding boxes vectors and the 3D hand keypoints vectors produced a single per-frame representation, and stacking across the temporal dimension yielded a tensor of size $150 \times \text{number of frames}$, capturing the full trajectory of hand motion throughout the video.

\paragraph{Global-Level Features}
We computed video-level statistics for both hands and tools based on the temporal changes of the bounding boxes' centers. Specifically, we calculated the following metrics: (1) still time - the proportion of frames in which an object's bounding box remains still; (2) average velocity and average acceleration - derived from the displacement of bounding box centers across consecutive frames. Averaged values were computed only during non-idle periods to reflect actual motion; (3) hands out of frame ratio - the proportion of frames in which a hand was not detected, indicating a large working volume; (4) the time needed to complete the task - video length; (5) autocorrelation between hands - the measure of how consistently hand movement patterns aligned with themselves across time.

\subsection{Proposed Framework}

\begin{figure}[!htb]
    \centering
    \includegraphics[width=\linewidth]{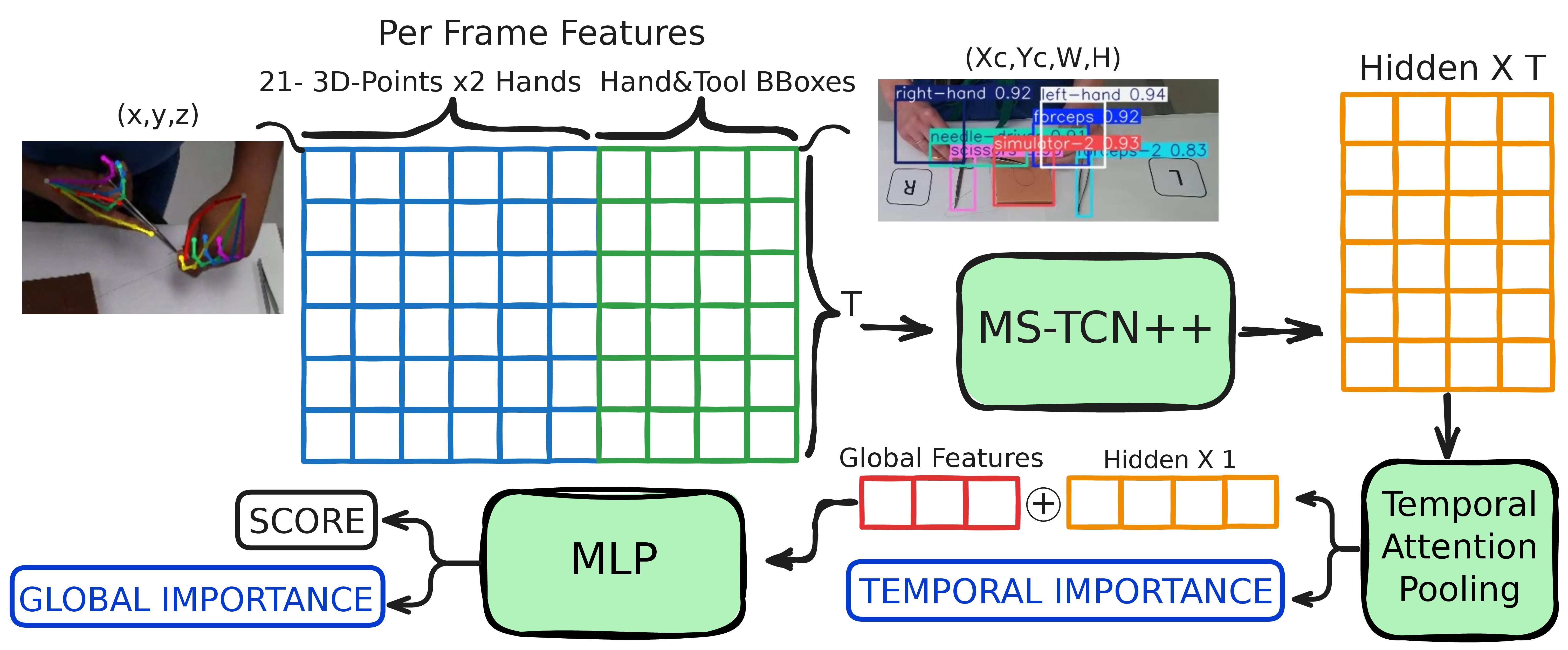}
    \caption{Overview of the proposed framework pipeline. The input video is processed into per-frame features consisting of 3D hand keypoints and bounding box information for hands and tools. Additionally, statistical global features are also calculated. The per-frame features are fed into the MS-TCN++ module, which generates a sequence of hidden representations over time. Temporal Attention Pooling is applied to extract a representative vector from the temporal sequence, while also producing temporal importance during inference. The temporally aggregated vector is concatenated with the global features and passed through an MLP to compute a final score. During inference, global importance is assessed using Shapley values for the entire video.}
    \label{fig:architecture}
\end{figure}

Our end-to-end framework is designed to generate skill predictions and provide interpretable feedback for trainees and educators. As illustrated in Figure~\ref{fig:architecture}, the pipeline consists of three main components: (1) a temporal backbone network (MS-TCN++) that encodes spatio-temporal dynamics from the per-frame features, (2) a temporal attention pooling module that highlights key video segments and compresses them into a compact representation, and (3) a Multi-Layer Perceptron (MLP) that fuses the temporal representation with the global statistical features to produce the final skill score. 
It is important to note that we preserve the temporal dimension, which enables the attention mechanism to identify when critical moments occur. In addition, we inject global statistical features only at the final fusion stage, thus maintaining their explanatory value for user feedback, rather than being diluted within intermediate learned representations. In the following paragraphs, we describe each component in detail.

\paragraph{Temporal Backbone}

The per-frame features, consisting of 3D hand keypoints and bounding boxes for hands and tools, are first organized into a feature matrix and passed into an MS-TCN++. With the classification head removed, the MS-TCN++ operates as a feature extractor. Each output vector at time step $t$ encodes temporal information of a consecutive frames window, resulting in a compact $F$-hidden dimension representation. Collectively, this yields a hidden representation of size $F\times T$, where $F$ is the feature hidden dimension and $T$ is the number of frames in the input sequence. Padding is applied to the feature matrix so that the temporal dimension is preserved, ensuring the output sequence length matches the input.

\paragraph{Temporal Attention Pooling.}
To capture long-range dependencies and identify salient moments, we apply a multi-head self-attention pooling mechanism on the hidden sequence. Given the spatio-temporal feature tensor $\mathbf{X} \in \mathbb{R}^{F \times T}$, we first permute it to $(T, F)$ to align with the standard attention formulation. Each frame representation is projected into query, key, and value embeddings using learnable linear layers. The resulting pooled representation is obtained as a weighted sum of frame features. During inference, this mechanism produces a frame-level importance map, which we later use to explain temporal saliency during inference 

\paragraph{Fusion with Global Features and MLP.}
The temporally pooled representation is concatenated with the global features vector we computed earlier. This fused representation is then passed through an MLP that outputs the final regression-based skill score. During inference, we also produce Global importance analysis as explained in the next section.

\section{Explainability}\label{Explainability}
Explainability is critical in surgical skill assessment, as it helps surgeons identify key procedural moments and motion features that most influence model predictions, supporting both learning and trust. In this work, we provide two complementary forms of explainability: temporal importance and global feature attribution.

\subsection{Temporal Importance}
In natural language processing, a well-known explainability approach is to attach a classification token ([CLS]) to the input and analyze its attention distribution across tokens: this highlights which words most strongly influenced the model’s decision \cite{devlin2019bert,perikos2024bert,chefer2021transformer}. Inspired by this paradigm, in our work, we consider video frames and their extracted features as analogous to words in an NLP sequence. As shown in Figure~\ref{fig:explainability_overview}-B, by applying interpretable attention pooling to these frame representations, we highlight temporal segments most critical to skill assessment, much like attention-based methods in NLP reveal influential tokens for classification.

\subsection{Global explainability}

We evaluated global explainability using SHAP (Shapley Additive exPlanations) values \cite{lundberg2017unified}, which quantify each feature’s contribution to the model’s prediction. Shapley values estimate the average marginal effect of a feature by comparing model outputs with and without that feature across feature subsets. We computed SHAP values for the extracted global features( statistical descriptors of interpretable video attributes) to analyze their influence on the predicted skill score.

As exact Shapley computation is intractable for complex models, we used the sampling-based SHAP approximation. Given the limited size of our task-specific datasets, we used the full dataset as the background distribution, yielding representative and stable attributions.

\section{Experiments}
\subsection{Score prediction}\label{score-prediction}
\paragraph{Training}
 Model training was conducted using an NVIDIA RTX A6000 GPU. A separate model was trained for each simulator .
For the loss function, we selected SORD\cite{Diaz_2019_CVPR} (Soft Ordinal Regression) Loss, which is specifically designed for ordinal regression. As a refinement of cross-entropy, it introduces ordinal penalties based on the distance from the true label, thereby encouraging predictions that are closer to the correct class.
\paragraph{Evaluation}
The dataset was split into training and test sets (70/30). Hyperparameters were selected via grid search with 4-fold cross-validation on the training set, using the average validation score for model selection. The final model was retrained on the full training set with optimal hyperparameters and evaluated on the held-out test set.

\paragraph{Results}
Table \ref{tab:simulator_metrics} presents the performance of the proposed framework across the three simulators. The \textit{Fascial Closure} task achieved the highest overall correlation with expert ratings ($r=0.778 \uparrow$) and the strongest coefficient of determination ($R^2=0.74 \uparrow$), indicating that the model explains most of the variance in this setting. The \textit{Knot Tying} task produced the lowest error values (RMSE $=1.276 \downarrow$) alongside a moderate correlation ($r=0.678 \uparrow$), suggesting stable predictions with reduced dispersion. In contrast, the \textit{Suturing} task yielded the smallest mean absolute error (MAE $=0.833 \downarrow$) but the lowest correlation ($r=0.622 \uparrow$) and $R^2=0.579 \uparrow$), reflecting greater variability and complexity in this procedure.


\begin{table}[ht]
\centering
\begin{tabular}{|l|c|c|c|c|c|}
\hline
\textbf{Simulator} & \textbf{Pearson Correlation ($r$) $\uparrow$} & \textbf{RMSE $\downarrow$} & \textbf{MAE $\downarrow$} & \textbf{R\textsuperscript{2}~$\uparrow$}  \\
\hline
Fascial Closure & 0.778 & 1.687 & 1.316 & 0.74 \\
Knot Tying & 0.678 & 1.276 & 0.982 & 0.649 \\
Suturing & 0.622 & 1.344 & 0.833 & 0.579 \\
\hline
\end{tabular}
\caption{Performance metrics for each simulator on its test set.}
\label{tab:simulator_metrics}
\end{table}


\subsection{Global Explainability Experiment}\label{globals-experiment}
\begin{figure}[!htb]
    \centering
    \includegraphics[width=0.8\linewidth]{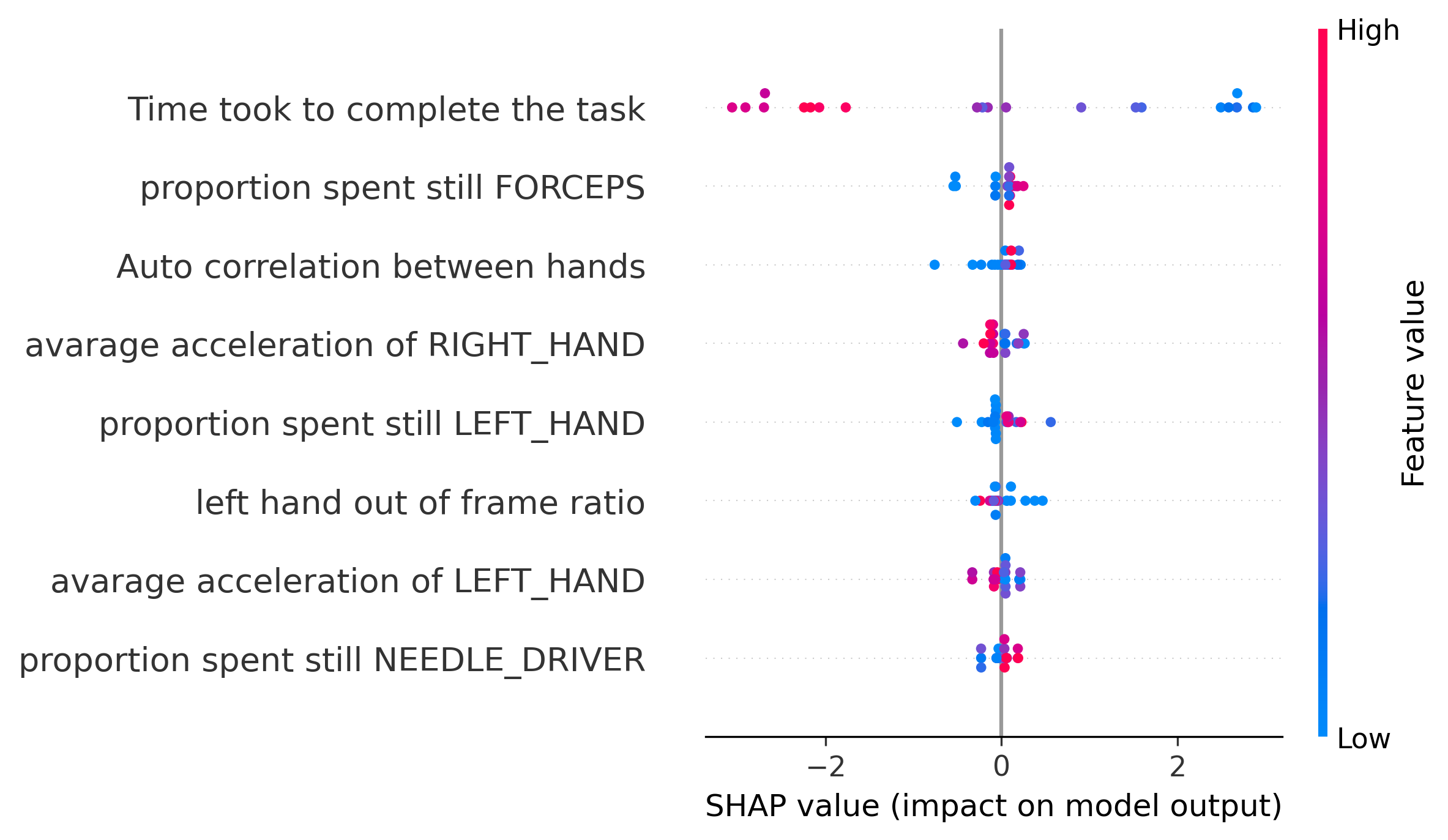}
    \caption{SHAP beeswarm plot for the \emph{Fascial Closure} test set. Each point of a feature corresponds to a video, with SHAP values (x-axis) indicating the feature’s impact on the predicted skill score and colors denoting feature values (blue = low, red = high). Task completion time, still forceps usage, and coordination between hands emerged as the most influential descriptors.}
    \label{fig:beeswarm}
\end{figure}

To validate our global explainability component, we computed SHAP values on the test set and created beeswarm plots for each simulator. For example, for the \emph{Fascial Closure} task, Figure~\ref{fig:beeswarm} shows Global importance across videos, with the horizontal axis representing SHAP values (i.e., the feature’s impact on the predicted skill score) and the color indicating the original feature value (blue = low, red = high). Features are ranked vertically by their overall importance.

Across simulators, the most influential features were:
\begin{itemize}
    \item \textbf{Fascial Closure:} \emph{time to complete the task} was the dominant feature, where shorter times were associated with higher predicted skill scores. Higher values of \emph{proportion of still forceps}, \emph{proportion of still left hand}, and \emph{autocorrelation between hands} contributed positively to predicted scores, while higher values of \emph{average acceleration} of both hands contributed negatively.
    \item \textbf{Knot Tying:} higher \emph{average speed of the right hand} and lower \emph{average acceleration of the right hand} were associated with higher predicted scores. Shorter \emph{completion time} also emerged as an influential factor.
    \item \textbf{Suturing:} Displayed patterns similar to Fascial Closure, but with a reversal in the dominance of features: higher values of \emph{proportion of time spent with still forceps} and shorter \emph{completion times} were more strongly associated with higher predicted scores, in that order of importance.

\end{itemize}

\section{Discussion and Conclusion}

Our results demonstrate that ExpOS enables accurate and interpretable surgical skill assessment while supporting feedback-oriented, autonomous training. Across three open-surgery tasks, the framework achieved strong correlation with expert ratings, indicating that data-driven temporal modeling of 3D hand motion captures clinically meaningful performance patterns.

A central contribution of this work is the shift from expert-defined evaluation criteria toward model-driven importance estimation. Rather than pre-specifying which metrics or segments determine skill, ExpOS learns discriminative temporal patterns directly from data and highlights the most informative moments through attention-based pooling. In parallel, the integration of global motion statistics enables feature-level attribution, identifying which interpretable characteristics most strongly influence predictions. This dual mechanism provides structured feedback at two levels: \textit{when} performance deviations occur and \textit{which} motion behaviors contribute to lower or higher skill scores.

Several limitations remain. The approach depends on reliable 3D hand reconstruction and clear visibility of both hands. Occlusions or challenging viewpoints may degrade performance. Future work will explore multimodal integration and validation on larger and more diverse surgical cohorts.

Overall, ExpOS demonstrates that combining weakly-supervised temporal importance learning with interpretable motion statistics enables scalable, explainable, and feedback-oriented surgical skill assessment.

%
%
%

\bibliographystyle{splncs04}
\bibliography{ExpOS-bibliography}

\end{document}